\documentclass[runningheads]{llncs}
\usepackage[T1]{fontenc}
\usepackage{verbatim}
\usepackage{graphicx}
\usepackage{xcolor} 
\usepackage[most]{tcolorbox}
\usepackage{booktabs}
\usepackage{caption}
\usepackage{colortbl}
\usepackage{multirow}
\usepackage{float}
\usepackage{verbatim}
\usepackage{subcaption} 
\usepackage{tikz}
\usetikzlibrary{shapes, arrows, positioning}

\usepackage[utf8]{inputenc}

\begin{document}
\title{ClaimPT: A Portuguese Dataset of Annotated Claims in News Articles}
%
%
\author{
Ricardo~Campos\inst{1,3}\orcidID{0000-0002-8767-8126}
Raquel~Sequeira\inst{1,3}\orcidID{0009-0004-4933-7568} \and
Sara Nerea\inst{1,3}\orcidID{0009-0002-2988-675X} \and
Inês Cantante\inst{2,3}\orcidID{0009-0002-3866-4550} \and
Diogo~Folques\inst{1,3}\orcidID{0009-0001-7580-8034} \and
Luís Filipe Cunha\inst{2,3}\orcidID{0000-0003-1365-0080} \and
João~Canavilhas\inst{1,7}\orcidID{0000-0002-2394-5264} \and
António Branco\inst{4, 5}\orcidID{0000-0002-7174-4942} \and
Alípio~Jorge\inst{2,3}\orcidID{0000-0002-5475-1382} \and
Sérgio~Nunes\inst{2,3}\orcidID{0000-0002-2693-988X} \and
Nuno~Guimarães\inst{2,3}\orcidID{0000-0003-2854-2891} \and
Purificação~Silvano\inst{2,3}\orcidID{0000-0001-8057-5338}
}
\authorrunning{R. Campos et al.}
%
\institute{
University of Beira Interior, Covilhã, Portugal \\
\email{\{ricardo.campos, raquel.sequeira, sara.n.silva, jc\}@ubi.pt}
\and
University of Porto, Porto, Portugal
\and
INESC TEC, Porto, Portugal 
\email{\{ines.cantante, diogo.folques, luis.f.cunha, alipio.jorge, sergio.nunes, nuno.r.guimaraes, purificacao.silvano\}@inesctec.pt}
\and
University of Lisbon, Lisbon, Portugal 
\email{antonio.branco@di.fc.ul.pt}
\and
NLX Group, Lisbon, Portugal \\
}


%
\maketitle              
\begin{abstract}
Fact-checking remains a demanding and time-consuming task, still largely dependent on manual verification and unable to match the rapid spread of misinformation online. This is particularly important because debunking false information typically takes longer to reach consumers than the misinformation itself; accelerating corrections through automation can therefore help counter it more effectively. Although many organizations perform manual fact-checking, this approach is difficult to scale given the growing volume of digital content. These limitations have motivated interest in automating fact-checking, where identifying claims is a crucial first step. However, progress has been uneven across languages, with English dominating due to abundant annotated data. Portuguese, like other languages, still lacks accessible, licensed datasets, limiting research, NLP developments and applications. In this paper, we introduce ClaimPT, a dataset of European Portuguese news articles annotated for factual claims, comprising 1,308 articles and 6,875 individual annotations. Unlike most existing resources based on social media or parliamentary transcripts, ClaimPT focuses on journalistic content, collected through a partnership with LUSA, the Portuguese News Agency. To ensure annotation quality, two trained annotators labeled each article, with a curator validating all annotations according to a newly proposed scheme. We also provide baseline models for claim detection, establishing initial benchmarks and enabling future NLP and IR applications. By releasing ClaimPT, we aim to advance research on low-resource fact-checking and enhance understanding of misinformation in news media.

\keywords{Claim Detection \and Fact-checking \and Misinformation  \and Annotated Dataset \and Low-resource Languages \and News Articles}
\end{abstract}
\setcounter{footnote}{0}
\section{Introduction}
Fact-checking plays a crucial role in mitigating misinformation, yet it remains a demanding, largely manual process that cannot keep pace with the scale and speed of misinformation \cite{inproceedings}. In recent years, organizations such as PolitiFact\footnote{\url{http://www.politifact.com}} have been established to perform systematic manual verification of claims through a multi-step process \cite{ijcai2021p619}: (i)~identifying claims; (ii)~prioritizing the most relevant ones; (iii)~collecting evidence from trustworthy sources; and (iv)~producing a verdict by comparing the claim against the gathered evidence. While effective for individual cases, this workflow becomes impractical when applied to the vast volume of daily online content. For instance, assessing a claim such as ``\emph{In just a few years, the number of immigrants has nearly quadrupled}'', requires retrieving information from multiple sources, evaluating reliability, and synthesizing evidence into a conclusion, an effort difficult to sustain at scale. To address these challenges, automated fact-checking has emerged as a promising research area \cite{guo-etal-2022-survey,PANCHENDRARAJAN2024100066}, with claim identification recognized as a crucial first step.

Initiatives such as CheckThat! at CLEF 2020 \cite{CheckThat} have encouraged research in this field. Yet, progress remains uneven across languages. Most corpora focus on English and on identifying claims in social media \cite{cheema-etal-2022-mm,mittal-etal-2023-lost} and political debates \cite{ivanov2024detectingcheckworthyclaimspolitical}, where language tends to be more informal, spontaneous, and conversational. Few studies have examined claims in news articles \cite{gangi-reddy-etal-2022-newsclaims}, a valuable yet underexplored domain. In contrast to other sources, news texts contain quotes from public figures embedded in the journalistic narrative, which often carry greater authority and shape public opinion. This underscores the need for dedicated resources that capture the unique properties of claims in news writing.

In this paper, we introduce ClaimPT, a novel dataset of 1,308 Portuguese news articles with annotated claims. Unlike existing corpora centered on social media or parliamentary discourse, ClaimPT is grounded in professionally edited journalistic content collected through a partnership with LUSA, the Portuguese News Agency. Building on the work of Reddy et al.~\cite{gangi-reddy-etal-2022-newsclaims}, who proposed NewsClaims for English COVID-19 news, ClaimPT extends coverage to a broader range of topics, reflecting the diversity of real-world reporting. In this work, a \textit{claim} is defined as a factual statement that asserts an alleged real-world fact of public interest, which can be verified \cite{inproceedings}. For the purpose of ClaimPT, we consider only declarative sentences occurring in the context of direct speech. Given a news article, our goal is to automatically identify the claim and its sub-elements. To ensure high-quality annotations, each article was independently labeled by two trained annotators, with the final version reviewed and consolidated by an experienced curator. The annotation scheme accommodates multiple labels per claim to improve claim contextualization. 

Finally, to assess how well this task can be approached automatically, we conduct baseline experiments using state-of-the-art pre-trained and generative language models, establishing initial benchmarks and highlighting key challenges for automated claim detection in Portuguese. Our contributions are thus threefold:
\begin{enumerate}
    \item We introduce ClaimPT, a new dataset of Portuguese news articles annotated for claim detection.
    \item We propose a new annotation scheme specifically tailored to capture the characteristics of claims in journalistic text.
    \item We conduct comprehensive baseline experiments on ClaimPT, leveraging state-of-the-art pre-trained and generative language models, providing benchmark results for automated claim detection in Portuguese.
\end{enumerate}
The remainder of this paper is structured as follows. Section \ref{sec:relatedwork} describes related contributions within the specific field of claim detection. Section \ref{sec:claimpt_annotation} describes the data acquisition process and the proposed annotation methodology, and provides a quantitative analysis including inter-annotator agreement scores. Section \ref{sec:claimpt_characterization} characterizes the ClaimPT dataset. Section \ref{sec:experiments} establishes baseline models for claim detection on the ClaimPT dataset, providing initial benchmarks. Finally, we present some final considerations in Section \ref{sec:conclusions}, pointing towards future work. 

\section{Related Work}\label{sec:relatedwork}
Fact-checking has become a crucial tool for verifying information of public interest disseminated by social actors \cite{repec:gam:jscscx:v:14:y:2025:i:9:p:514-:d:1733231}. In recent years, automated fact-checking has been commonly framed as a pipeline comprising several subtasks, including claim detection (identifying check-worthy claims), and claim verification (assessing veracity) \cite{inproceedings,vandermeer2025hintsoftruthmultimodalcheckworthinessdetection,PANCHENDRARAJAN2024100066,schlichtkrull2023averitecdatasetrealworldclaim,thorne-etal-2018-fever}. Early research on claim detection explored opinionated statements in online discussions such as weblogs and Wikipedia forums \cite{6337079}. The ClaimBuster system \cite{inproceedings} formalized the task by ranking statements from U.S. presidential debates according to their check-worthiness. Subsequent initiatives such as the CLEF CheckThat! lab series \cite{atanasova2018overviewclef2018checkthatlab,CheckThat,elsayed2021overviewclef2019checkthatautomatic} consolidated the task through shared benchmarks, primarily focused on political debates and social media content. These works typically frame claim detection as a classification or ranking problem that separates factual from non-factual statements \cite{vasileva-etal-2019-takes}. Despite steady progress, supervised models lag behind human performance \cite{ijcai2021p619}, motivating the expansion of CheckThat! to new domains such as COVID-19 misinformation. Recent work has broadened the scope and modalities of claim detection, applying it to diverse domains including health \cite{alam2020call2arms,shaar-etal-2021-findings}, environmental issues \cite{stammbachenvironmental}, political debates \cite{elsayed2021overviewclef2019checkthatautomatic,gencheva-etal-2017-context,inproceedings}, and social media posts \cite{10.1007/978-3-031-13643-6_29,shaar-etal-2021-findings}. Other approaches rely on fact-checking archives \cite{chowdhury-etal-2025-fact5,automatedfact}, synthetic claim generation \cite{vandermeer2025hintsoftruthmultimodalcheckworthinessdetection}, or multimodal settings that integrate text, video, and image data \cite{giedemann2025viclaimmultilingualmultilabeldataset,zlatkova-etal-2019-fact}. Nevertheless, textual claims remain central to professional fact-checking workflows \cite{ijcai2021p619}.

Among text-based resources, NewsClaims \cite{gangi-reddy-etal-2022-newsclaims} represents an important step forward, providing manually annotated English news claims with additional attributes, such as span, claimer, topic, and stance, that enrich contextual understanding. Such attribute-aware formulations support fact-checkers by clarifying who made the claim and what it concerns, facilitating interpretation and verification \cite{ijcai2021p619}.
Despite these advances, multilingual coverage remains limited. Most resources are English-centric, with few available for Portuguese. Datasets like Fake.Br \cite{fakenewspropor} offer partial coverage for Brazilian and European Portuguese, but remain narrow in scope. "Polígrafo", a Portuguese fact-checking outlet, was used to label short statements for X-Fact, yet large-scale, open Portuguese resources are still lacking. As noted by Panchendrarajan and Zubiaga \cite{PANCHENDRARAJAN2024100066}, this scarcity hampers the development of localized fact-checking tools. While multilingual and cross-lingual models mitigate the gap to some extent \cite{ijcai2021p619}, native datasets remain crucial to capture language-specific nuances. More recently, systems such as Explainable Automatic Fact-Checking for Journalists \cite{ijcai2025p1140} have introduced end-to-end automated fact-checking frameworks designed to assist journalists. Although developed in English, they underscore the need for scalable, language-specific datasets to extend support to other languages, including Portuguese. To address these limitations, ClaimPT focuses on European Portuguese and on claims appearing in professionally edited news articles, aiming to advance claim detection in low-resource settings and the development of computational tools.

\section{The ClaimPT Dataset}\label{sec:claimpt_annotation}

\subsection{Data Acquisition} 

A protocol established with the Portuguese news agency LUSA granted access to a corpus of news articles spanning politics, society, economics, international affairs, sports, health, culture, science and environment, local news, and technology. The initial retrieval yielded 1,808 articles, from which a pilot sample of 100 was analyzed to refine the selection criteria. This exploratory analysis revealed that politics and international news contained the highest density of claims, whereas sports exhibited minimal claim content. Accordingly, a second request prioritized claim-rich topics to enhance the dataset’s representativeness. Since most claims occurred in direct speech, the retrieval parameters were further refined to target sentences containing reporting verbs (e.g., "said", "declared"). Data acquisition proceeded in two sequential phases to ensure temporal consistency, covering news articles from 4 January 2022 to 28 December 2023. The final corpus comprises 1,308 news articles obtained through this iterative process.

\subsection{Annotation scheme and process} 
The ClaimPT annotation scheme was designed to systematically identify and categorize claims and non-claims in Portuguese news articles, thereby contributing to the creation of resources for automated claim detection and fact-checking. It builds upon established frameworks such as NewsClaims~\cite{gangi-reddy-etal-2022-newsclaims} and ClaimBuster~\cite{inproceedings}. The former served as the main structural foundation, from which we adopted key dimensions including topic, stance, claim object, and claimer. The latter provided the conceptual distinction between claims and non-claims and offered valuable insights into criteria to identify them. 

\begin{figure}
      \centering
      \includegraphics[width=\linewidth]{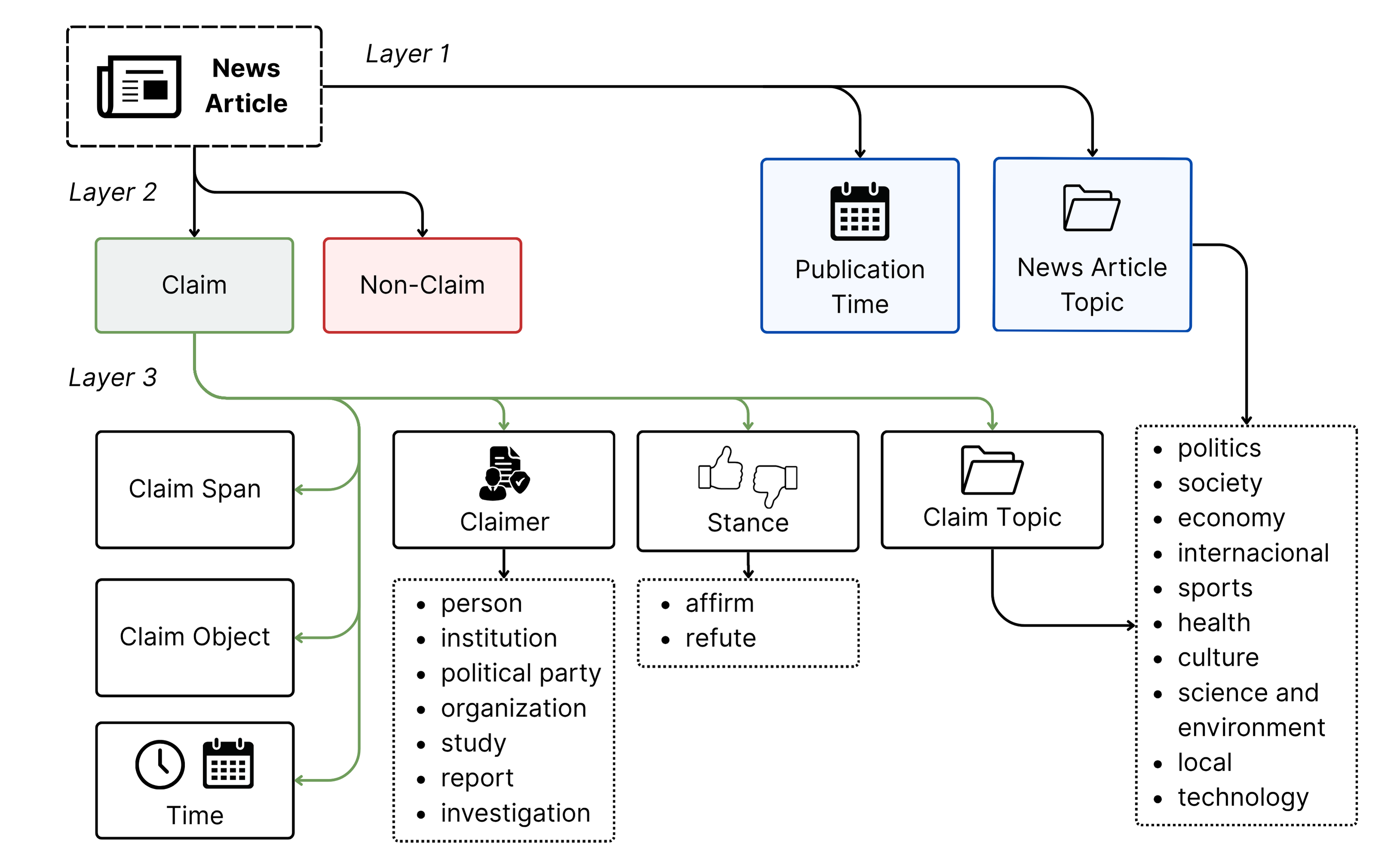}
\caption{Overview of the annotation scheme.}
\label{fig:annotation-scheme}
\end{figure}

Building on this foundation, ClaimPT advances previous schemas by introducing a multi-layer architecture that connects entity-level information with document metadata and relational links. In contrast to earlier approaches that treated each claim as an isolated unit, ClaimPT integrates metadata, entities with attributes, and link structures, thereby enabling a more comprehensive representation of the claim and of contextual information. This integration allows annotators to capture complex instances in which, for example, identifying a claim requires considering the broader text or retrieving information about the claimer from preceding discourse. This contextual information is captured, for instance, through the annotation of metadata and through an \textit{Identity\_link} which represents the correferential connections regarding the claimer. Alongside the dimensions of topic, stance, claim span, claim object, and claimer adopted from NewsClaims, our scheme introduces an additional temporal dimension, a label representing the time interval during which the claim was made. This temporal information is crucial not only for assessing the claim’s relevance to current events but also for defining the appropriate timeframe for verification, which must precede the claim’s date of utterance. In addition, our scheme enriches the representation of the claimer entity by assigning it an attribute (e.g., person, institution). This addition allows for a more fine-grained characterization of the claim and supports analyses that account for how the nature of the source \cite{10.1145/3770077} influences the framing, credibility, and dissemination of claims. 
Figure~\ref{fig:annotation-scheme} gives an overview of the annotation scheme, organized in three main layers. 

\begin{figure}
      \centering
      \includegraphics[width=\linewidth]{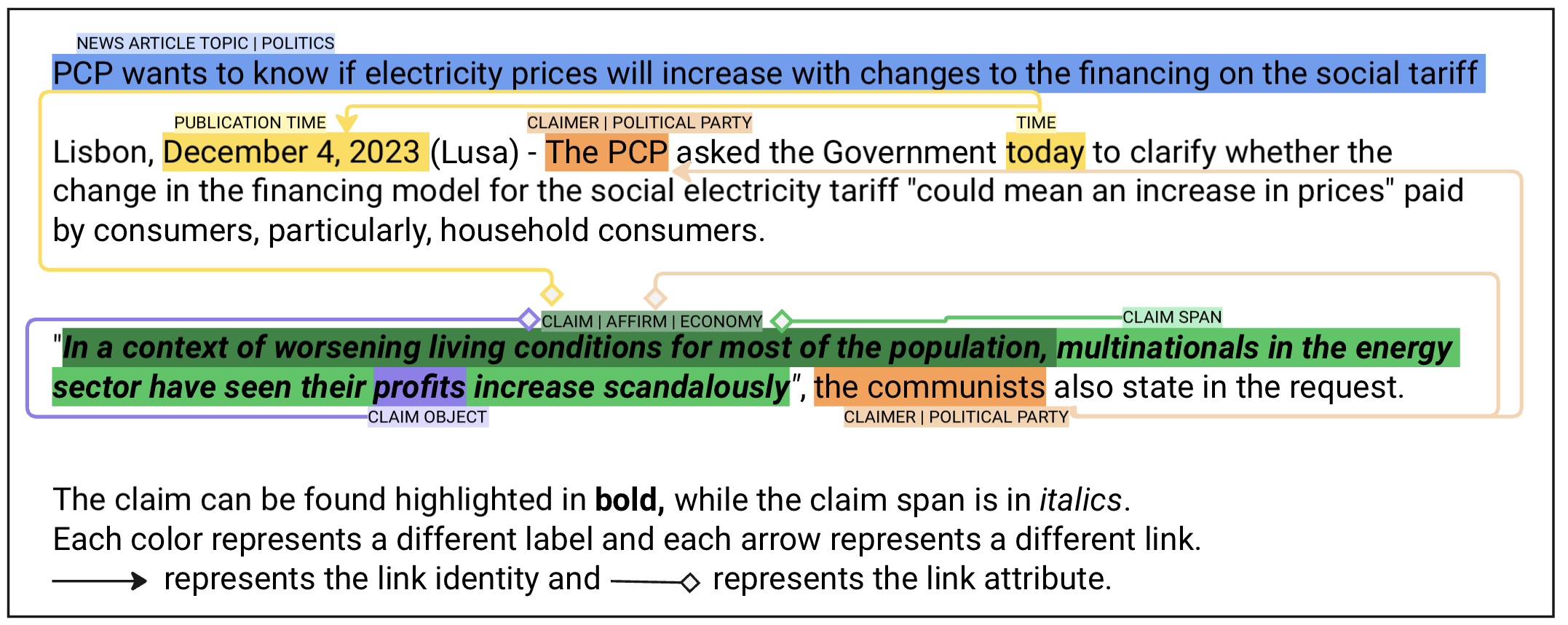}
       \caption{Annotation Scheme example.}
       \label{fig:example}
   \end{figure}

The \textbf{first layer} encodes the \textit{metadata} of each news article, capturing its thematic classification and publication date under the labels \textit{News Article Topic} and \textit{Publication Time}. The first assigns each article to one of ten predefined categories: \textit{Politics}, \textit{Society}, \textit{Economy}, \textit{International}, \textit{Sports}, \textit{Health}, \textit{Culture}, \textit{Science and Environment}, \textit{Technology}, or \textit{Local}. The latter records the publication time of the article’s release date. At the core of the scheme is the distinction between \textit{claims} and \textit{non-claims}, the \textbf{second layer}. \textit{Claims} are defined as verifiable factual statements of public interest, typically expressed in declarative form and attributed to an external source rather than the journalist (\textit{"The Government had promised to extend the surface metro to Costa da Caparica and that promise was not fulfilled"}). In contrast, \textit{non-claims} correspond to subjective or speculative utterances, such as opinions, beliefs, or predictions, which cannot be verified against factual evidence (\textit{"I believe that the decision of the President of the Republic [of Guinea-Bissau] is not right."}). In a \textbf{third layer}, each claim is further decomposed into a set of fine-grained components capturing its internal structure. The \textit{claim span} identifies the textual segment that conveys a complete factual assertion, while the \textit{claim object} denotes the specific target or proposition being asserted. The \textit{claimer} represents the agent responsible for the statement, and has attributes such as a person, organization, institution, or report. Only non-official sources, that is, those not considered inherently unquestionable according to journalistic standards, are annotated as claimers in our framework. Identifying the claimer is therefore crucial for establishing whether a statement qualifies as a claim, as utterances from institutional authorities such as the WHO are often exempt from explicit verification. Temporal information explicitly or implicitly associated with the claim is marked under the \textit{time} label, and the \textit{stance} label specifies whether the claimer affirms or refutes the statement. Finally, each claim is assigned a \textit{claim topic}, drawn from the same taxonomy as the article-level topic but determined solely by the claim’s immediate content. Additionally, we include two link structures: \textit{attribute} and \textit{identity} links. The first type connects claims with their corresponding span, object, claimer, and time, ensuring the internal coherence of each annotation. The second type establishes referential connections across the text, linking pronouns or abbreviated mentions to their full antecedents, such as “he” to “Prime Minister António Costa”, which is also novelty in relation to prior proposals. Figure \ref{fig:example} illustrates an example of the annotation as proposed by the ClaimPT framework. 

The annotation methodology followed a multi-stage, iterative process designed to ensure accuracy, consistency, and reproducibility. Before large-scale annotation began, a structured training phase was implemented to assess the annotators’ understanding of the scheme. Each annotator completed two trial rounds on a sample of ten articles, after which results were discussed with the curator to resolve disagreements and refine label definitions and attribute specifications. This preliminary phase also provided an estimate of the average annotation time-approximately twenty minutes per article, and an expected productivity of around fifty articles per week. The annotation process was conducted over a six-month period, from March to September 2025.

Based on the outcomes of this pilot phase, both the annotation guidelines and methodology were reviewed and finalized prior to the large-scale annotation. The process was then organized into four stages. First, annotators carefully read the full text of each news article. Next, they annotated the metadata, namely the \textit{News Article Topic} and \textit{Publication Time}. This was followed by claim identification, during which annotators examined all direct-speech sentences to determine whether they constituted claims or non-claims, applying the definitional and linguistic criteria outlined in the manual, which is made available on the resource GitHub\footnote{https://github.com/LIAAD/ClaimPT}. In this phase, all components of each claim were also marked. For the annotation of the \textit{claimer} label, our guidelines draw on a journalistic classification of claim sources. Official sources presented as claimers are considered non-verifiable by journalists, as they represent entities from which professionals typically obtain factual information \cite{gradim2000manual,schmitz2011classification}. Consequently, statements uttered by such claimers are regarded as trustworthy and are not annotated. In contrast, statements whose claimers may raise uncertainty or require further scrutiny are annotated, reflecting the verification processes that would occur in real-world journalistic practice. 

Each news article was independently annotated by two trained annotators with academic backgrounds in communication studies. The annotations were then validated by an experienced curator with expertise in linguistics and a proven record in the curation of other annotation initiatives \cite{nunes-etal-2024-text2story}. The entire annotation process was supervised by a senior researcher with extensive experience in large-scale annotation projects (e.g., \cite{piskorski-etal-2025-semeval}). The supervisor organized regular meetings in which annotators and curator discussed cases of disagreement, reviewed challenging examples, and, in instances where consensus could not be reached, the final decision on the annotation was made by the supervisor to ensure consistency and methodological coherence across the dataset. All annotation work was carried out using the INCEpTION platform \cite{klie-etal-2018-inception}. The resulting dataset is available online in the GitHub repository with a persistent DOI \cite{claimpt2025}.

\subsection{Inter-Annotator Agreement}

Following the annotation process and curator review, agreement was evaluated at the span level through a one-to-one match between annotators, applying a \textbf{±4-character} boundary tolerance to mitigate minor selection discrepancies. Two levels of analysis were conducted: (i) agreement on \textit{Claim} vs.~\textit{Non-Claim} labels, and (ii) agreement on claim attributes: \textit{Claim span}, \textit{Claim object}, \textit{Claimer}, and \textit{Time}, restricted to cases where both annotators agreed on the claim span label.

Table~\ref{tab:IAA} presents the results, displaying matched spans and disagreements. While nominal Krippendorff’s Alpha is a standard IAA metric, it assumes a fixed, shared item set, which is unsuitable for span-based annotation where one annotator may omit a span. Moreover, cases labeled as \texttt{<no annotation>} distort results if included and obscure disagreement if excluded. To address these issues, we adopted \textbf{overlap-based metrics}, specifically per-label Jaccard (\% agreement) and inter-annotator F1 (\textit{Sørensen–Dice}), which more accurately capture agreement at the span level.

\begin{table}[t]
\centering
\caption{Inter-Annotator Agreement regarding Claim/Non-claim as well as the Claim attributes (only on the subset where the annotators agreed on the Claim).}
\label{tab:IAA}
\begin{subtable}[t]{0.4\textwidth}
\centering
\caption{Claim/Non-claim}
\label{tab:IAA_claim_nonclaim}
\resizebox{\textwidth}{!}{%
\begin{tabular}{|l|cc|}
\hline
\textbf{\begin{tabular}[c]{@{}l@{}}Labels/\\ Metrics\end{tabular}} & \multicolumn{1}{l|}{\textbf{Claim}} & \multicolumn{1}{l|}{\textbf{Non-Claim}} \\ \hline
\textbf{F1}                                                        & \multicolumn{1}{c|}{0.5036}         & 0.5435                                  \\ \hline
\textbf{Jaccard(\%)}                                               & \multicolumn{1}{c|}{33.65}          & 37.32                                   \\ \hline
\rowcolor[HTML]{EFEFEF} 
\textbf{Micro-F1}                                                  & \multicolumn{2}{c|}{\cellcolor[HTML]{EFEFEF}0.5390}                           \\ \hline
\rowcolor[HTML]{EFEFEF} 
\textbf{Macro-F1}                                                  & \multicolumn{2}{c|}{\cellcolor[HTML]{EFEFEF}0.5236}                           \\ \hline
\end{tabular}%
}
\end{subtable}
\hfill
\begin{subtable}[t]{0.48\textwidth}
\centering
\caption{Claim attributes}
\label{tab:IAA_claim_attribtues}
\resizebox{\textwidth}{!}{%
\begin{tabular}{|l|cccc|}
\hline
\textbf{\begin{tabular}[c]{@{}l@{}}Attributes/\\ Metrics\end{tabular}} & \multicolumn{1}{l|}{\textbf{Span}} & \multicolumn{1}{l|}{\textbf{Object}} & \multicolumn{1}{l|}{\textbf{Claimer}} & \multicolumn{1}{l|}{\textbf{Time}} \\ \hline
\textbf{F1}                                                            & \multicolumn{1}{c|}{0.7593}          & \multicolumn{1}{c|}{0.4393}        & \multicolumn{1}{c|}{0.7980}           & 0.9348                             \\ \hline
\textbf{Jaccard(\%)}                                                   & \multicolumn{1}{c|}{61.20}           & \multicolumn{1}{c|}{28.15}         & \multicolumn{1}{c|}{66.39}            & 87.77                              \\ \hline
\rowcolor[HTML]{EFEFEF} 
\textbf{Micro-F1}                                                      & \multicolumn{4}{c|}{\cellcolor[HTML]{EFEFEF}0.7196}                                                                                                    \\ \hline
\rowcolor[HTML]{EFEFEF} 
\textbf{Macro-F1}                                                      & \multicolumn{4}{c|}{\cellcolor[HTML]{EFEFEF}0.7329}                                                                                                    \\ \hline
\end{tabular}%
}
\end{subtable}
\end{table}

For \textit{Claim/Non-Claim} (Table~\ref{tab:IAA_claim_nonclaim}), agreement is moderate, with a Jaccard score of 33.65\% and F1 of 0.5036 for \textit{Claim}. While these results may seem modest, they primarily reflect the intrinsic ambiguity and subjectivity of the task. The annotation guidelines specify a credibility criterion, whereby statements from inherently reliable sources (e.g., the World Health Organization) should remain unannotated. However, this criterion was applied inconsistently in some cases, resulting in documents annotated by only one annotator and consequently reducing span-overlap scores. This issue, identified during the curation phase, was subsequently corrected. Restricting the evaluation to spans annotated by both annotators improves agreement, with Jaccard rising to 36.71\% and F1 to 0.5371. These findings are consistent with prior work, where reported agreement levels for claim identification typically range between 0.3 and 0.5 (30–50\%)~\cite{gangi-reddy-etal-2022-newsclaims}. 
Among claim attributes (Table~\ref{tab:IAA_claim_attribtues}), \textit{Claim object} exhibits the lowest agreement (F1 = 0.4393; Jaccard = 28.15\%), reflecting its interpretive nature. In contrast, \textit{Claim span}, \textit{Claimer}, and \textit{Time} achieve substantially higher agreement (F1 between 0.7593 and 0.9348; Jaccard 61–88\%), indicating that once a claim is identified, its main attributes can be reliably annotated.


\section{ClaimPT dataset characterization}\label{sec:claimpt_characterization}

The \textit{ClaimPT} dataset consists of 1,308 news articles provided by the \textit{Lusa News Agency}, with an average length of 542.8 words per article. Of these, 1,090 articles are annotated with \textit{Claims} and \textit{Non-Claims}, while the remaining 218 include only metadata annotations, specifically \textit{News Article Topic} and \textit{Publication Time}. The number of annotated \textit{News Article Topics} and \textit{Publication Times} is equal to the total number of documents (1,308), as each document contains exactly one annotation of each type. 

In terms of thematic coverage of the news articles, the most frequent topic is \textit{International} news, accounting for 21.79\% of the dataset, followed by \textit{Politics} with 13.84\%. A complete distribution of documents by topic is presented in Table~\ref{tab:topic_table}.

\begin{table}[t]
\caption{Distribution of topics regarding the news article and the individual claims in the \textit{ClaimPT} dataset.}
  \centering
\begin{subtable}[t]{0.41\textwidth}
\subcaption{News Article Topics}
\label{tab:topic_table}
\centering
\resizebox{\textwidth}{!}{%
\begin{tabular}{lcc}
\hline
\multicolumn{1}{c}{\textbf{News Topic}} & \textbf{Count} & \textbf{Percent (\%)} \\ \hline
international                           & 285            & 21.79                 \\
politics                                & 181            & 13.84                 \\
health                                  & 154            & 11.77                 \\
local                                   & 145            & 11.09                 \\
society                                 & 129            & 9.86                  \\
sports                                  & 108            & 8.26                  \\
science and environment                 & 100            & 7.65                  \\
economy                                 & 100             & 7.65                  \\
culture                                 & 65             & 4.97                  \\
technology                              & 41             & 3.13                  \\ \hline
\textbf{Total}                          & \textbf{1308}  & \textbf{100}       \\ \hline
\end{tabular}%
}
\end{subtable}
\hfill
\begin{subtable}[t]{0.45\textwidth}
\subcaption{Claim Topics}
\label{tab:claim_topic_distribution}
\centering
\resizebox{\textwidth}{!}{%
\begin{tabular}{lcc}
\hline
\multicolumn{1}{c}{\textbf{Claim Topic}} & \textbf{Count} & \textbf{Percentage (\%)} \\ \hline
Politics                                 & 105            & 22.68                     \\
Society                                  & 100             & 21.60                     \\
International                            & 73             & 15.77                     \\
Economy                                  & 47             & 10.15                      \\
Local                                    & 43             & 9.29                     \\
Health                                   & 43             & 9.29                      \\
Science and Environment                  & 37             & 7.99                     \\
Culture                                  & 9             & 1.94                      \\
Technology                               & 3             & 0.65                      \\
Sports                                   & 3              & 0.65                      \\ \hline
\textbf{Total}                           & \textbf{463}   & \textbf{100}           \\ \hline
\end{tabular}%
}
\end{subtable}
\end{table}
  
Of the 1,090 fully annotated documents, 273 ($\approx$ 25\%) contain both \textit{Claim} and \textit{Non-Claim} annotations, while the remaining documents include only \textit{Non-Claims}. In total, 463 \textit{Claims} were identified, corresponding to an average of 0.42 per document, and 4,393 \textit{Non-Claims}, averaging 4.03 per document. This yields a global distribution of 9.53\% \textit{Claims} and 90.47\% \textit{Non-Claims}. This distribution is in line with the nature and editorial principles of news articles, which favor factual information over the explicit formulation of check-worthy claims. As a result, only a limited subset of statements in news articles qualifies as claims under our annotation criteria, making a low claim frequency an expected characteristic of this genre. At the same time, this scarcity highlights the intrinsic difficulty of the claim identification task in news articles: relevant claims are sparse, embedded, and often indirect. However, their limited number does not diminish the significance of the task; rather, it underscores its relevance and serves to expose a core challenge that automated systems must address when operating on real-world news data. As a matter of fact, although it results in a class imbalance, it provides a realistic setting for training and evaluating claim detection models, encouraging the development of systems that are robust to sparse positive instances and capable of distinguishing check-worthy claims from the predominant background of non-claim content.

Table~\ref{tab:claim_topic_distribution} presents the distribution of annotated \textit{Claim} topics within the \textit{ClaimPT} dataset. Overall, the distribution highlights a strong prevalence of politically and socially oriented claims, consistent with the news agency’s editorial focus and the broader relevance of these topics in public discourse. 

Among the 670 annotated \textit{Claimers}, 563 were classified as \textit{person} and 66 as \textit{organization}, with the remaining labels retaining only a marginal count. The number of \textit{Claimer} annotations exceeds the total number of \textit{Claims} (463) because, in cases where accurate identification required it, two textual segments were annotated to represent a single claimer within the same claim. With respect to \textit{Stance} associated with each \textit{Claim}, a total of 463 annotations were recorded, comprising 446 instances labeled as "affirm" and 17 as "refute". Finally, \textit{Claim spans} and \textit{Claim objects} amount to 523 and 551 annotations, respectively, revealing only a few cases where more than one claim span and object were annotated per claim. 
The ClaimPT dataset was split into \textit{train} and \textit{test} sets according to several considerations. The standard 80\%/20\% split was applied at document-level. Additionally, documents were carefully selected so that the ratio properties closely match those of the original dataset. Both \textit{train} and \textit{test} maintain the same \textit{Claim:Non-Claim} ratio as the full set (1:9.48), and the total counts for each claim attribute are also distributed approximately according to the 80\%/20\% split. 

\section{Task and Baselines}\label{sec:experiments}
We evaluate the potential of the ClaimPT dataset by framing claim detection as a span classification task: given a text \textit{t}, the model predicts a set of triples $(b,e,c)$, where $b$ and $e$ indicate the start and end of a span, and $c \in {\text{Claim},\text{Non-Claim}}$ denotes the class. To facilitate future research, we also establish baseline models and standard evaluation metrics.

\subsection{Baselines}

We consider two types of LLM-based baselines: encoder models (BERT-style) and generative models using few-shot learning. For the generative baselines, we use two Gemini models \cite{geminiteam2024geminifamilyhighlycapable}: Gemini-2.5-Flash and Gemini-2.5-Flash-Lite. These models offer large context windows and strong general capabilities, serving as a challenging baseline. Structured outputs are extracted via the LangExtract~\cite{langextract} library, using a one-shot System 2 Attention prompt \cite{weston20232attentionisneed} made available at our repository (translated to English for better comprehension).
For the encoder baseline, claim extraction is formulated as a token classification task. Each token $w_i$ in a sequence $W = [w_1, ..., w_n]$ is labeled as part of a claim or non-claim: $\mathcal{L} = {\text{B-Claim, I-Claim, B-Non-Claim, I-Non-Claim, O}}$. We fine-tune BERTimbau \cite{10.1007/978-3-030-61377-8_28} on ClaimPT. To handle documents exceeding BERT's 512-token limit, we use two strategies: (1) Sentence-level segmentation, where each sentence is processed independently, preserving boundaries, which are important for claim detection; (2) Chunking with overlap (stride), where documents are split into 512-token chunks with 128-token overlap to capture cross-boundary context. For overlapping tokens, only the prediction from the first occurrence is retained.

\subsection{Evaluation}
The ClaimPT corpus provides ground truth annotations defined by span offsets and their corresponding labels. For evaluation, we apply consistent span-level metrics across both generative and BERT-based models. In the case of BERT, token-level outputs are first aggregated into spans that mark the start and end boundaries of each predicted claim or non-claim. These predicted spans are then compared to the ground truth, with a match considered correct only if both boundaries and the label exactly align. Model performance is assessed using precision, recall, and F1-score at the span level. Precision measures the proportion of correctly predicted spans among all predictions, recall quantifies the proportion of ground-truth spans successfully identified, and F1-score represents their harmonic mean, providing a balanced view of overall effectiveness.

\subsection{Results and Discussion}
Results (Table~\ref{tab:results}) show that encoder-based models outperform generative baselines. Gemini 2.5 Flash Lite achieves an F1 score of 22.52\%, while Gemini 2.5 Flash achieves an F1 score of 36.50\%. These results are aligned with the expected since that although these models have the capabilities to achieve high effectiveness across a wide range of tasks \cite{geminiteam2024geminifamilyhighlycapable}, previous studies also demonstrate their difficulty in extracting span-based annotations \cite{gptstructme}. Among encoder baselines, BERT with sentence segmentation performs best, reaching F1-scores of 30.57 for claims and 69.27 for non-claims.

\begin{table}[t]
\centering
\caption{Claim extraction results for generative and BERT models on the ClaimPT dataset. BERT-Sent refers to the sentence-level segmentation strategy, while BERT-Chunk refers to the chunking strategy with overlap (stride).}
\label{tab:results}
\begin{tabular}{lcccc}
\hline
\textbf{Model} & \textbf{Label} & \textbf{Precision (\%)} & \textbf{Recall (\%)} & \textbf{F1 (\%)} \\
\hline
\multirow{3}{*}{Gemini Flash Lite (Generative)} 
& Claim & 2.81 & 58.06 & 5.35 \\
& Non-Claim & 28.94 & 50.57 & 36.81 \\
& Micro Avg & 14.43 & 51.28 & 22.52 \\
\hline
\multirow{3}{*}{Gemini 2.5 (Generative)} 
& Claim & 6.68 & 61.29 & 12.05 \\
& Non-Claim & 33.03 & 70.07 & 44.90 \\
& Micro Avg & 24.78 & 69.23 & 36.50 \\
\hline
\multirow{3}{*}{BERT-Chunk} 
& Claim & 40.38 & 22.58 & 28.97 \\
& Non-Claim & 55.96 & 68.71 & 61.68 \\
& Micro Avg & 55.24 & 64.31 & 59.43 \\
\hline
\multirow{3}{*}{BERT-Sent} 
& Claim & 37.50 & 25.81 & 30.57 \\
& Non-Claim & 63.35 & 76.42 & 69.27 \\
& Micro Avg & 61.88 & 71.59 & 66.38 \\
\hline
\end{tabular}
\end{table}
A closer inspection reveals a misalignment between the annotation guidelines and the generative model’s interpretation of what constitutes a claim. The model frequently identifies journalistic quotations as claims - although such quotes may contain verifiable content and public relevance, they are not statements made by social agents, as required by the guidelines. Owing to surface-level similarities, the generative models also tend to conflate unverifiable generalizations with legitimate claims. Concerning the direct citation rule, the model often extends its predictions beyond quoted text, incorporating adjacent context in an attempt to form a complete sentence. Despite these inconsistencies, the model demonstrates a partial ability to separate verifiable statements from opinions. For instance, in one case it correctly annotates a claim segment while labeling three opinions within the same sentence as non-claims. According to the ground truth, however, the entire sentence should be annotated as a claim, since the guidelines do not allow overlapping claim and non-claim segments. Nonetheless, this behavior suggests that the model captures a meaningful distinction between factual and opinionated content, effectively decomposing complex sentences into verifiable and non-verifiable parts.


Despite this, both baselines show relatively low effectiveness compared to previous studies \cite{prabhakar-etal-2020-claim}.  This difference is partly explained by the task formulation: while some works treat claim detection as a multi-class classification problem, our setup adopts a span classification perspective, where the model must not only distinguish claims from non-claims but also identify their precise textual boundaries. This makes the task inherently more complex—particularly when dealing with long documents that exceed encoder context windows. Nonetheless, this formulation offers greater practical value for real-world applications, as claim boundaries often depend on the surrounding context for accurate interpretation. In settings such as newsroom fact-checking, span-based systems therefore provide a more interpretable and context-aware framework than simple classification approaches. In summary, encoder-based baselines, especially BERT models with sentence segmentation, constitute a solid foundation for future research on this dataset. While model performance is not the main focus of this resource paper, these baselines establish a meaningful reference point for subsequent studies to refine and extend. 

\section{Conclusions and Future work}\label{sec:conclusions}

In this paper, we introduce ClaimPT, a new European Portuguese dataset for claim detection. It comprises 1,308 news articles annotated at the span level with claims, non-claims, and additional claim components. This resource enables a fine-grained approach to claim detection, supporting the development of systems that identify claims directly within news text. Beyond the dataset, we provide detailed annotation guidelines, inter-annotator agreement analysis, dataset characterization, and baseline experiments framing claim detection as a span classification task with corresponding evaluation metrics. Together, these elements form a solid foundation for advancing research in claim detection. Future directions include (1) extending the dataset to a multilingual context using the provided guidelines, (2) improving baselines with state-of-the-art models such as ModernBERT \cite{warner2024smarterbetterfasterlonger}, and (3) leveraging annotated claim attributes to build more explainable, structured models, deepening understanding of how claims are expressed and contextualized in text.

\section{Limitations}
\textbf{Corpus}. The dataset comprises European Portuguese news articles spanning nine distinct topics. While it is currently limited to this language, its design and annotation guidelines can be readily extended to other languages. Although topic distribution is not uniform and the number of non-claims considerably exceed that of claims (463 vs. 4,393), this imbalance mirrors the natural frequency of verifiable statements in real news content and thus constitutes a valuable characteristic rather than a limitation, providing a realistic setting for assessing claim extraction systems under conditions that reflect the true distribution of claims in journalistic text, an essential aspect for models intended for real-world application. Despite detailed guidelines and rigorous quality control, annotators and curators, though experts, may still introduce minor biases or subjective interpretations inherent to the task. \textbf{Baseline Models}. For the commercial baselines, we employed two LLMs from the Gemini family. These models may evolve or become deprecated over time, potentially affecting reproducibility. In addition, minimal prompt engineering was applied, which may limit performance; however, since these models primarily serve as baselines and the paper’s main contribution lies in the dataset itself, this setup is considered sufficient.

\subsubsection*{Preprint and Version of Record}
This preprint has not undergone peer review (when applicable) or any post-submission improvements or corrections. The Version of Record of this contribution is published in {Advances in Information Retrieval. ECIR 2026. Lecture Notes in Computer Science, vol 16486 Springer, Cham.}, and is available online at https://doi.org/10.1007/978-3-032-21321-1\_58.

\subsubsection*{Acknowledgments}
This work is funded by national funds through FCT – Fundação para a Ciência e a Tecnologia, I.P., under the support UID/50014/2025 (https://doi.org/ 10.54499/UID/50014/2025).
Luís Filipe Cunha thanks the Fundação para a Ciência e Tecnologia (FCT), Portugal for the Ph.D. Grant (2024.042 02.BD).
António Branco, Raquel Sequeira, Sara Nerea, Diogo Folques would like to acknowledge the project ACCELERAT.AI - Multilingual Intelligent Contact Centers, funded by the covid-recovery program PRR-Plano de Recuperação e Resiliência, through IAPMEI (C625734525-00462629); PORTULAN CLARIN - Research Infrastructure for the Science and Technology of Language, funded by LISBOA2030 (FEDER-01316900); hey, Hal, curb your hallucination!, funded by FCT-Fundação para a Ciência e Tecnologia (2024.07592.IACDC). Ricardo Campos, Alípio Jorge, Nuno Guimarães and Purificação Silvano would also like to acknowledge project StorySense, with reference 2022.09312.PTDC (DOI 10.54499/2022.09312.PTDC). João Canavilhas would also like to acknowledge project Obiajor, with reference 2023.18007.ICDT. We would also like to acknowledge Rodrigo Silva for preparing Figure 1 and the Portuguese News Agency (LUSA) for granting access to the 1,308 news articles.

%
%
\bibliographystyle{splncs04}
\bibliography{mybibliography}
\end{document}